\documentclass[default]{sn-jnl}

\usepackage{amsmath,amssymb,amsfonts}%
\usepackage{textcomp}%
\usepackage{mathtools}
\usepackage{comment}
\usepackage{tabularx}
\usepackage{array}
\usepackage{booktabs}%
\usepackage{subcaption}
\usepackage{multirow}
\usepackage{enumitem}
\usepackage{hyperref}
\usepackage[labelsep=space]{caption}
\usepackage{multirow}%
\usepackage{todonotes}

\usepackage{amsthm}%
\usepackage{mathrsfs}%
\usepackage[title]{appendix}%
\usepackage{xcolor}%
\usepackage{manyfoot}%
\usepackage{algorithm}%
\usepackage{algorithmicx}%
\usepackage{algpseudocode}%
\usepackage{listings}%
\usepackage{graphicx}%

\usepackage{adjustbox}

\newcolumntype{P}[1]{>{\centering\arraybackslash}p{#1}}
\newcolumntype{C}[1]{>{\arraybackslash}p{#1}}

\def\BibTeX{{\rm B\kern-.05em{\sc i\kern-.025em b}\kern-.08em T\kern-.1667em\lower.7ex\hbox{E}\kern-.125emX}}

\usepackage{verbatim}

\begin{document}

\title[Towards Interpretable Motion-level Skill Assessment in Robotic Surgery]{Towards Interpretable Motion-level Skill Assessment in Robotic Surgery}


\author*[1]{\fnm{Kay} \sur{Hutchinson}}\email{kch4fk@virginia.edu}

\author[2]{\fnm{Katherina} \sur{Chen}}\email{kyc3rk@virginia.edu}

\author[1]{\fnm{Homa} \sur{Alemzadeh}}\email{ha4d@virginia.edu}

\affil*[1]{\orgdiv{Department of Electrical and Computer Engineering}, \orgname{University of Virginia}, \orgaddress{\city{Charlottesville}, \postcode{22903}, \state{VA}, \country{USA}}}

\affil[2]{\orgdiv{Department of Urology}, \orgname{University of Virginia}, \orgaddress{\city{Charlottesville}, \postcode{22903}, \state{VA}, \country{USA}}}


\abstract{
\textbf{Purpose:} We study the relationship between surgical gestures and motion primitives in dry-lab surgical exercises towards a deeper understanding of surgical activity at fine-grained levels and interpretable feedback in skill assessment.

\textbf{Methods:} We analyze the motion primitive sequences of gestures in the JIGSAWS dataset and identify inverse motion primitives in those sequences. Inverse motion primitives are defined as sequential actions on the same object by the same tool that effectively negate each other. We also examine the correlation between surgical skills (measured by GRS scores) and the number and total durations of inverse motion primitives in the dry-lab trials of Suturing, Needle Passing, and Knot Tying tasks.

\textbf{Results:} We find that the sequence of motion primitives used to perform gestures can help detect labeling errors in surgical gestures. Inverse motion primitives 
are often used as recovery actions to correct the position or orientation of objects or may be indicative of other issues such as with depth perception. The number and total durations of inverse motion primitives in trials are also strongly correlated with lower GRS scores in the Suturing and Knot Tying tasks.

\textbf{Conclusion:} The sequence and pattern of motion primitives could be used to provide interpretable feedback in surgical skill assessment. Combined with an action recognition model, the explainability of automated skill assessment can be improved by showing video clips of the inverse motion primitives of inefficient or problematic movements.

}




\maketitle



\section{Introduction}
\label{sec:introduction}

Surgical robotic systems and simulators provide automated scoring and feedback on training exercises based on system log, video, and kinematic data. Existing training tools and simulators mainly focus on efficiency, safety, and task/procedure-specific metrics \cite{hutchinson2022analysis}, but assess overall performance during a demonstration and are not descriptive enough to pinpoint inefficiencies or errors. To improve explainability, previous works performed skill assessment at different levels of the surgical hierarchy where surgical tasks can be decomposed into smaller units of activity called gestures and motions such as motion primitives (MPs) \cite{hutchinson2023compass} or dexemes \cite{varadarajan2009data, van2021gesture}. 
Leveraging lower-level activities can improve the performance of multi-granularity models \cite{valderrama2022towards}
and the task-generalization \cite{hutchinson2023evaluating} of activity recognition models. However, the relationship between units of activity at different levels of the surgical hierarchy is relatively unexplored. Surgical tasks have been modeled with graphs using gestures \cite{ahmidi2017dataset}, MPs \cite{hutchinson2023compass}, and dexemes \cite{varadarajan2009data}. \cite{zhao2022murphy} trained a model for step, task, and action recognition that learned inter- and intra-relations between units of activity at different levels of the hierarchy. But the relationship between gestures and MPs has not been well studied. 

Many works have analyzed gestures for skill assessment as an indication of expertise.
For example, some studies found that experts use fewer gestures, make fewer errors, and have more predictable transitions~\cite{vedula2016analysis}. 
Or some gestures are more indicative of skill than others~\cite{hutchinson2022analysis, reiley2009task, varadarajan2009data, cao1996task, koskinen2022movement, inouye2022assessing}.
Skill assessment can be also augmented by the emerging sub-field of error detection \cite{van2021gesture} since there is a correlation between skill level and errors \cite{inouye2022assessing, hutchinson2022analysis}.
Towards this, \cite{yasar2020real} and \cite{li2022runtime} trained models to detect errors, and \cite{hutchinson2022analysis} defined and analyzed gesture-level procedural and executional errors. \cite{anastasiou2023keep} trained a contrastive regression transformer for skill assessment that could also detect those errors. 
However, there is a large variability in the correct performances of gestures due to style or expertise and it is impossible to define all errors or incorrect performances of gestures, limiting the effectiveness of supervised machine learning models trained on small datasets.

Skill assessment has also been performed at the motion-level using data-driven methods. \cite{cao1996task} and \cite{uemura2016procedural} found differences between experts and novices in the numbers and durations of actions performed during laparoscopic tasks. 
\cite{koskinen2022movement} examined efficiency in microsurgical suturing based on time and the cosine similarity of movement patterns. But, these works 
were limited to statistical analyses of the motions without considering patterns in the motion sequences. 
Models such as Hidden Markov Models (HMMs) have also been used for motion-level skill assessment where the motions were modeled with hidden states and learned from data \cite{varadarajan2009data}. 
\cite{varadarajan2009data} showed that the sequence of states (motions) in a gesture-level HMM could be indicative of skill level. 
Interestingly, \cite{reiley2009task} found that gesture-level HMMs for skill assessment were better than task-level models, and that building task-level models with gestures as the states was better than learning the states from unlabeled kinematic data. This motivates a lower, motion-level analysis of surgical activity.
Gestures have also been modeled as strings comprised of finer-grained activities mapped to an alphabet \cite{forestier2018surgical} 
or recognized using bottom-up approaches that first learn finer-grained motions~\cite{despinoy2015unsupervised}. 
However, motions learned from data in an unsupervised manner are not easily human interpretable and do not correspond with labeled units of surgical activity \cite{reiley2009task}.

Recent efforts have been directed at increasing the interpretability of surgical skill assessment.
Several methods use metrics such as time, path length, and economy of motion, as inputs to their models so that the outputs are explainable and objective \cite{brown2020bring, khalid2022or} 
while others identify or localize the parts of the trial that contribute to the predicted score by visualizing trajectories or features \cite{forestier2018surgical, fawaz2019accurate}. 
\cite{wang2020towards} proposed a model for gesture recognition and skill assessment with running intermediate scores to identify problematic gestures. But, providing gesture-specific feedback based on identifying patterns in finer-grained units of surgical activity has not been explored yet.

In this study, we take a step towards improving skill assessment with interpretable feedback in robot-assisted surgical training tasks by examining finer-grained motions and their relationship to gestures and expertise level. We make the following contributions:

\begin{itemize}
    \item Analysis of the relationship between surgical gestures and finer-grained MPs which shows that the sequence of MPs can help detect labeling errors in gestures.
    
    \item Definition and detection of \textit{inverse motion primitives} in the execution of gestures and tasks. Inverse MPs are often used as recovery actions to correct the position or orientation of objects, or may be indicators of other issues such as poor depth perception.  
    
    \item Analysis of the types, frequencies, and durations of inverse MPs across gesture types and skill levels which shows that they strongly correlate with lower GRS scores, and could be used to improve the interpretability of skill assessment by identifying the specific motions within a trial that contribute to lower scores.
\end{itemize}

\section{Methods}
\label{sec:methods}

This section presents an overview of the surgical process model and datasets used in this study, and our methods for the analysis of motion primitive sequences and their relationship to gestures and surgical skill.


\begin{figure}
    \centering
    \includegraphics[page=1, trim=0.5in 0.5in 6in 0.5in, clip, width=\textwidth]{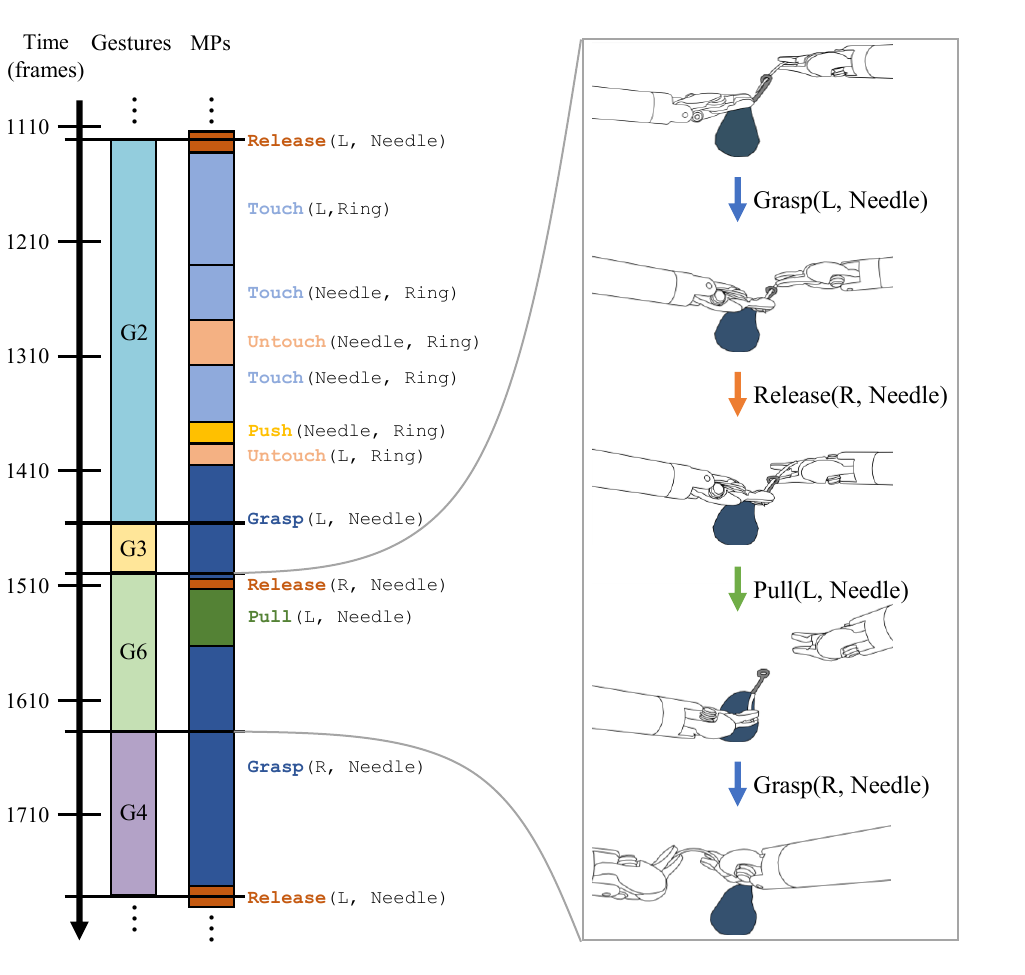}
    \caption{Example of gesture and motion primitive labels for a trial of Needle Passing and the motion primitive sequence extracted for G6.}
    \label{fig:mp_seq_extraction}
\end{figure}

\subsection{Surgical Process Model}
Surgical procedures can be decomposed into finer levels of granularity,
including phases, steps, tasks, gestures, and actions \cite{lalys2014surgical}. 
Surgical \textit{tasks} such as suturing and knot tying are modeled as a sequence of \textit{gestures} (shown as grammar graphs \cite{ahmidi2017dataset}) that represent units of surgical activity with semantic meaning and specific intents \cite{hutchinson2023towards}. Gestures (also called sub-tasks or surgemes) can be decomposed into finer-grained atomic motions, called \textit{motion primitives} (MPs)~\cite{hutchinson2023compass}, dexemes \cite{varadarajan2009data, van2021gesture}, or action triplets~\cite{nwoye2022rendezvous}, representing generalizable units of activity that enable lower-level analyses \cite{hutchinson2023compass}. 
Motions are often modeled as triplets \cite{hutchinson2023compass, nwoye2022rendezvous} 
encoding the type of action, the tool used, and the object affected. 
We use the MP model and notation from \cite{hutchinson2023compass} (e.g., Grasp(L, Needle)) with the left and right tools abbreviated as ``L'' and ``R''.

\subsection{Datasets}
We use the publicly available JIGSAWS dataset \cite{gao2014jhu} containing kinematic and video data from dry-lab simulation training experiments, with 39 trials of Suturing, 28 trials of Needle Passing, and 36 trials of Knot Tying. These trials were performed by two expert surgeons ($>$100 hours of robotic surgical experience), two intermediate surgeons (10-100 hours), and four novice surgeons ($<$10 hours). Each trial was annotated with gesture labels and a global rating score (GRS) based on a modified OSATS approach, comprised of subscores such as for Respect for tissue, Suture/needle handling, and Flow of operation. 
The COMPASS dataset \cite{hutchinson2023compass} provides MP labels for the tasks in the JIGSAWS dataset. In this analysis, we focus on gestures that occur multiple times during a trial as listed in Table \ref{tab:gestures_and_mps}.

\begin{table}[h!]
\caption{JIGSAWS gestures \cite{gao2014jhu} and surgeon-defined MP sequences \cite{hutchinson2023compass} (N = Needle).}
\label{tab:gestures_and_mps}
\begin{tabular}{p{0.5cm} p{2.7cm} c l}
\toprule
\multicolumn{2}{c}{Gesture and Description} & Task & Surgeon-defined Motion Primitive Sequence \\
\midrule
\multirow{2}{*}{G2} & \multirow{2}{3cm}{Positioning needle} & S & Touch(Needle, Fabric) \\
 & & NP & Release(L, N), Touch(Needle, Ring), Push(Needle, Ring) \\
\multirow{2}{*}{G3} & \multirow{2}{3cm}{Pushing needle through tissue} & S & Touch(Needle, Fabric), Push(R, N), Grasp(L, N) \\
 & & NP & Grasp(L, N) \\
\multirow{2}{*}{G4} & \multirow{2}{3cm}{Transferring needle from left to right} & S & Grasp(R, N), Release(L, N) \\
 & & NP & Grasp(R, N) \\    
\multirow{2}{*}{G6} & \multirow{2}{3cm}{Pulling suture with left hand} & S & Grasp(L, N), Release(R, N), Pull(L, N) \\
 & & NP & Release(R, N), Pull(L, N) \\
\multirow{2}{*}{G8} & \multirow{2}{3cm}{Orienting needle} & S & Grasp(L, N), Release(R, N), Grasp(R, N), Release(L, N) \\
 & & NP & Release(R, N), Grasp(R, N) \\
\multirow{2}{*}{G12} & \multirow{2}{3cm}{Reaching for needle with left hand} & \multirow{2}{*}{KT} & \multirow{2}{*}{Grasp(L, Thread), Release(R, Thread)} \\
 & & & \\
\multirow{2}{*}{G13} & \multirow{2}{3cm}{Making C loop around right hand} & \multirow{2}{*}{KT} & \multirow{2}{*}{Pull(L, Thread), Touch(R, Thread)} \\
 & & & \\
\multirow{2}{*}{G14} & \multirow{2}{3cm}{Reaching for suture with right hand} & \multirow{2}{*}{KT} & \multirow{2}{*}{Grasp(R, Thread)} \\
 & & & \\
\multirow{2}{*}{G15} & \multirow{2}{3cm}{Pulling suture with both hands} & \multirow{2}{*}{KT} & \multirow{2}{*}{Pull(L, Thread) Pull(R, Thread)} \\
 & & & \\
\bottomrule
\end{tabular}%
\end{table}

\subsection{MP Sequence Analysis}

We first extracted the video data for each gesture trial in JIGSAWS, then two expert robotic surgeons from urology and gynecology defined the ideal sequences of MPs by reviewing the gesture definitions and videos of gesture trials (see Table \ref{tab:gestures_and_mps}). 
The surgeons omitted Touch and Untouch MPs immediately preceding or following Grasp and Release MPs, respectively, of the same object with the same tool, so we combine MPs in the transcripts meeting those requirements prior to our MP sequence extraction and analysis.

We extracted the corresponding MP sequence for each gesture trial from the COMPASS dataset based on the start and end frames of the gestures in JIGSAWS. If an MP overlapped at the beginning or end of a gesture, it was \textit{included in the MP sequence for the gesture}. For example, in Figure \ref{fig:mp_seq_extraction}, Grasp(L, Needle) is included in the MP sequences for G2, G3, and G6. But this also means that in the video clip for the gesture, \textit{parts of those MPs get cut off in the video} (because the start and end frames of the gesture are used to create the video clips).



We plot the frequencies of the MP sequences for each gesture in each task to examine their variety, and create state graphs to visualize their transitions.
This provides insights into how surgeons perform gestures including specific patterns in the MPs, and identifies inconsistencies in the gesture annotations.  

\subsection{Inverse MP Definition and Analysis}

The analysis of MP sequences shows some easily identifiable patterns in the execution of gestures across different tasks and trials.
One of the most common patterns are two or more sequential MPs performed by the same tool on the same object whose verbs effectively negate or undo each other (e.g., Touch(Needle, Fabric), Untouch(Needle, Fabric), Touch(Needle, Fabric) in G2 as shown in Figure \ref{fig:mp_seq_extraction}), which we define as \textit{Inverse MPs}. 
Inverse MPs most often are symptoms of inefficient or erroneous execution of gestures, but they could also be necessary in some cases (e.g., the ``Release(R, Needle), Grasp(R, Needle)'' in G8). For each gesture, we only focus on the inverse MPs that are not part of the surgeon-defined sequences in Table \ref{tab:gestures_and_mps}.
Video clips of each gesture containing one or more instances of inverse MPs are manually reviewed to confirm their presence and gain insight about the relationship between their types and the gestures they occur in.

\subsection{Inverse MPs and Skill Level}
To further assess inverse MPs as indicators of inefficiencies in the performance of surgical gestures and tasks, we examine their correlation with surgeon expertise level and GRS scores. 
For each trial, we count the number of inverse MPs and their total durations. This summation includes inverse MPs in all gestures in a trial even if it was not considered in the analysis above (e.g., G1), and was not subject to manual review to remove instances when the inverse MP was not seen. We then analyze the frequency of the inverse MPs across expertise levels and calculate Spearman's correlation ($ \rho $) between the frequency and durations of inverse MPs and the GRS scores or sub-scores.

\section{Results}
\label{sec:results}

\subsection{MP Sequence Analysis}
\begin{figure}[t!]
    \centering
    \includegraphics[trim = 2.75in 0in 0.in 0in, clip, width=\textwidth]{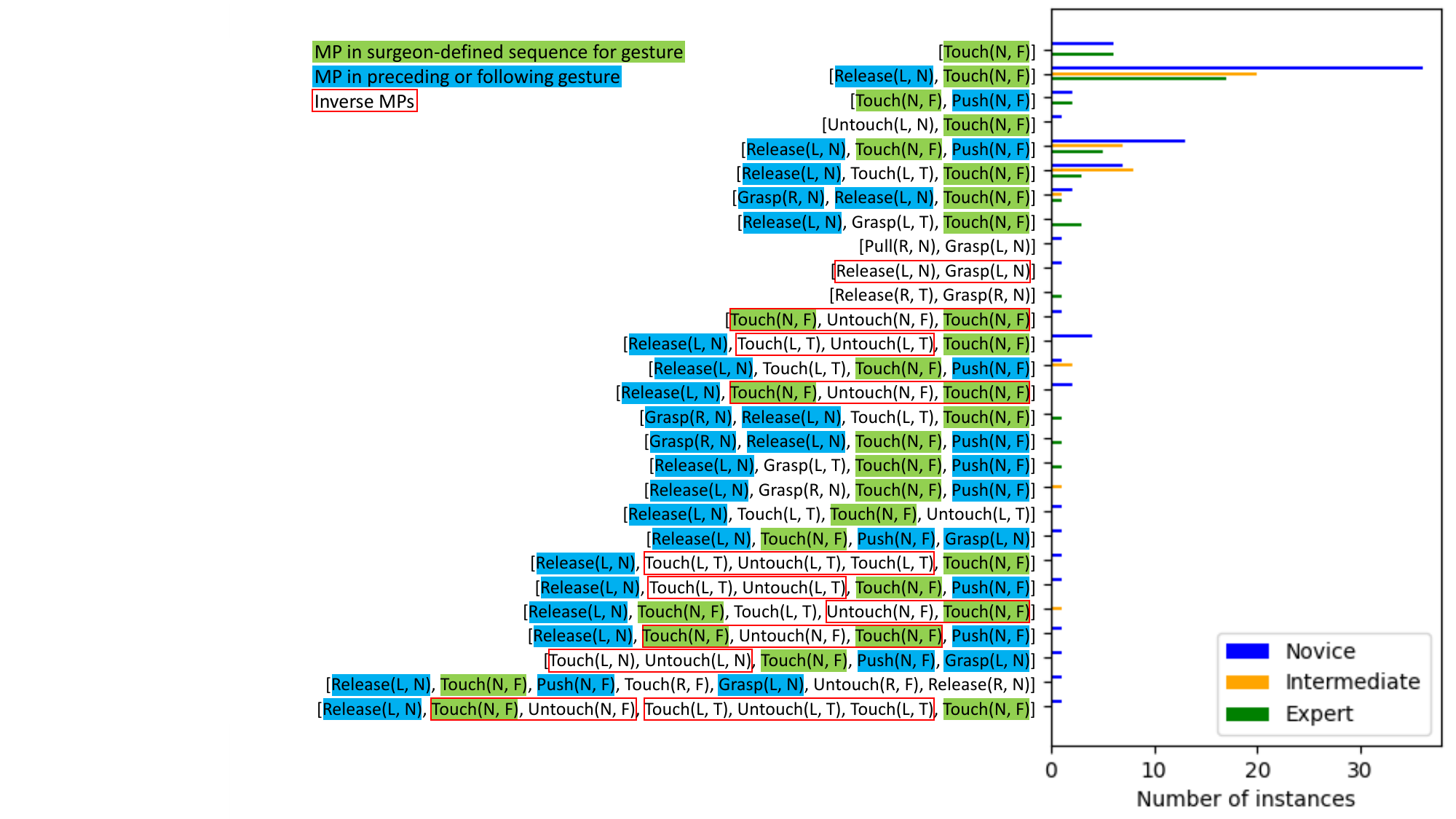}
    \caption{Number of instances of each MP sequence in G2 in Suturing (F = Fabric, N = Needle, and T = Thread). Figure best viewed in color.}
    \label{fig:expNIE_hist}
\end{figure}

For each gesture, we compare the variety of MP sequences to the surgeon-defined sequences in Table \ref{tab:gestures_and_mps}.
As an example, the vertical axis of Figure \ref{fig:expNIE_hist} lists all of the different MP sequences used to perform G2 in Suturing. 
Many of the MP sequences contained strings of MPs that effectively negated each other, as highlighted in the red boxes in Figure \ref{fig:expNIE_hist}, so we identify them as inverse MPs and conduct further analysis on them in the following sections.

\subsubsection{MPs at the Boundaries of Gestures}

Several gestures contain MPs from the surgeon-defined MP sequence of gestures that are before or after them according to gesture grammar graphs in \cite{hutchinson2023compass, ahmidi2017dataset} as listed in Table \ref{tab:mps_at_boundaries}.
For Needle Passing, the surgeon-defined MP sequence of G2 includes Push(Needle, Fabric), so there were fewer boundary issues with MPs from gestures before and after G2 compared to Suturing. 
Partial or full repetitions of the surgeon-defined sequences were present in all gestures except G6 and G13 and could indicate that the gesture was attempted multiple times.
Future work should explore if these boundary inconsistencies contribute to the poor performance of machine learning models at the transitions between gestures.

\begin{table}[h!]
    \vspace{-1.5em}
    \centering
    \caption{Common MPs from neighboring gestures found in each gesture.}
    \label{tab:mps_at_boundaries}
    \begin{tabular}{c c l c}
        \toprule
        Task & Gesture & MPs from the Preceding or Following Gesture \\
        \toprule
        \multirow{4}{*}{Suturing} & G2 & Release(L, Needle) from the G8 before  \\
         & & and/or the Push(Needle, Fabric) from the G3 after \\
         & G3 & Release(R, Needle) of the G6 after  \\
         & G6 & Grasp(R, Needle) of the G4 after  \\
        \midrule
        \multirow{5}{*}{Needle Passing} & G3 & Release(R, Needle) in the G6 after \\
         & G4 & Release(L, Needle) in the G2 after  \\
         & G6 & Grasp(L, Needle) from the G3 before  \\
         & & and/or the Grasp(R, Needle) from the G4 after  \\
         & G8 & Release(L, Needle) from the G2 after  \\
         \midrule
        \multirow{3}{*}{Knot Tying} & G12 & Pull(L, Thread) from the G13 after  \\
         & G14 & Pull(L, Thread) and Pull(R, Thread) from the G15 after  \\
         & G15 & Release(L, Thread) and Release(R, Thread) from the G11 after  \\
        \bottomrule
    \end{tabular}
    \vspace{-0.5em}
\end{table}

Analysis of MP sequences can help with identifying gesture annotation errors. \cite{van2020multi} listed twelve amendments to the Suturing gesture labels, where four
contain MPs from the gesture immediately after the original gesture and two do not contain all of the MPs in the labeled gesture. 
For example, we found that in the second trial of Suturing performed by subject C, the MP sequence for the G6 at frames 1506-1685 is Release(R, Needle), Pull(L, Needle), Grasp(R, Needle), Release(L, Needle). 
The third and fourth MPs belong to G4 according to Table \ref{tab:gestures_and_mps}, which matches the correction in \cite{van2020multi}.
Inverse MPs can also indicate gesture annotation errors since all gesture clips from Needle Passing containing ``Touch/Untouch(Needle, Ring)'' that were labeled G4, G6, and G8 
were actually G2.

\begin{table}[t!]
    \centering
    \caption{Number of inverse MPs in each gesture (F/R = Fabric or Ring).} 
    \label{tab:inv_pairs}
   
    \resizebox{\textwidth}{!}{
    \begin{tabular}{l|ccccccccc|c}
    \toprule
        Inverse MP & G2 & G3 & G4 & G6 & G8 & G12 & G13 & G14 & G15 & Total \\
        \midrule
        Touch(L, Needle) Untouch(L, Needle) & 1 & 10 & 0 & 2 & 3 & - & - & - & - & 16 \\
        Touch(R, Needle) Untouch(R, Needle) & 1 & 0 & 7 & 2 & 4 & - & - & - & - & 14 \\
        Grasp(L, Needle) Release(L, Needle) & 2 & 8 & 2 & 6 & 12 & - & - & - & - & 30 \\
        Grasp(R, Needle) Release(R, Needle) & 3 & 6 & 3 & 1 & 2 & - & - & - & - & 15 \\
        Touch(L, Thread) Untouch(L, Thread) & 7 & 3 & 0 & 2 & 0 & 5 & 0 & 0 & 2 & 19 \\
        Touch(R, Thread) Untouch(R, Thread) & 0 & 0 & 1 & 8 & 0 & 3 & 0 & 15 & 8 & 35 \\
        Grasp(L, Thread) Release(L, Thread) & 0 & 0 & 0 & 1 & 0 & 17 & 1 & 3 & 9 & 31 \\
        Grasp(R, Thread) Release(R, Thread) & 0 & 0 & 0 & 0 & 0 & 6 & 0 & 2 & 13 & 21 \\
        Touch(L, F/R) Untouch(L, F/R) & 3 & 2 & 0 & 0 & 0 & - & - & - & - & 5 \\
        Touch(R, F/R) Untouch(R, F/R) & 0 & 0 & 1 & 7 & 0 & - & - & - & - & 8 \\
        Grasp(L, F/R) Release(L, F/R) & 3 & 0 & 0 & 0 & 0 & - & - & - & - & 3 \\
        Grasp(R, F/R) Release(R, F/R) & 0 & 0 & 0 & 0 & 0 & - & - & - & - & 0 \\
        Touch(Needle, F/R) Untouch(Needle, F/R) & 49 & 2 & 2 & 2 & 1 & - & - & - & - & 56 \\
        Push(Needle, F/R) Pull(R, Needle) & 0 & 4 & 0 & 0 & 0 & - & - & - & - & 4 \\
        \midrule
        Total & 69 & 35 & 16 & 31 & 22 & 31 & 1 & 20 & 32 & 257 \\
    \bottomrule
    \end{tabular}
    }
\end{table}

\subsection{Analysis of Inverse MPs}

The number of inverse MPs by type (after subtracting instances not seen in the manual review) is shown in Table \ref{tab:inv_pairs}. The number of gestures containing one or more inverse MPs by gesture and expertise level are, respectively, shown in Tables \ref{tab:inv_pairs_gesture_clips_gestures} and \ref{tab:inv_pairs_gesture_clips_NIE}. 

\subsubsection{Categories of Inverse MPs}

Our manual review found that inverse MPs could be broadly 
categorized as relating to depth perception issues, poor positioning or orientation of the needle or suture in the graspers, or incidental actions to adjust the environment.

\textbf{Depth perception} issues displayed overshoot in the movement towards an object, running into an object when reaching for another object, or multiple attempts to grab it. 
This appears as ``Touch/Untouch(L, Needle)'' in G3 while reaching for the needle with the left tool after it emerged from the fabric or ring, or in G14 which had the most instances of ``Touch/Untouch(R, Thread)'' in Table \ref{tab:inv_pairs} where surgeons struggled to grasp the suture on the left side of the knot due to poor depth perception or occlusion by one of the tools.

\begin{figure}[h!]
    \centering
    \includegraphics[page=2, trim=0.25in 2.8in 6.25in 2.7in, clip, width=\textwidth]{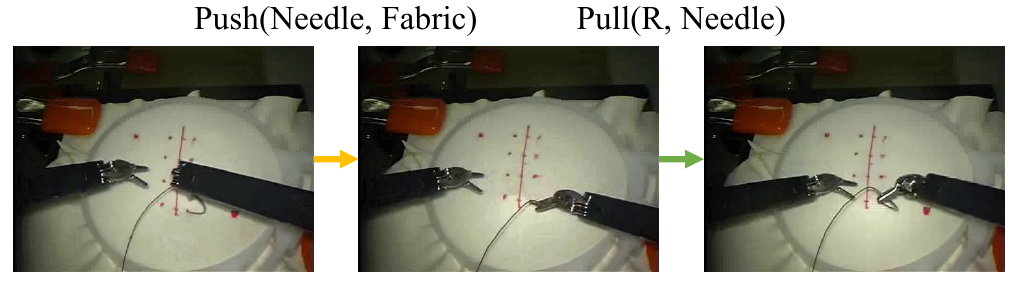}
    \caption{Example of a Push(Needle, Fabric) Pull(R, Needle) inverse MP in Suturing.}
    \label{fig:inv_pair_err}
    \vspace{-1em}
\end{figure}

\textbf{Poor positioning or orientation of the needle or suture} was a significant contributor to inverse MPs. In these cases, inverse MPs were used as recovery actions to fix the position or orientation of the needle or suture from previous gestures.
In suturing, the needle should be held perpendicular to its curve and two-thirds from its tip, and inserted with the tip perpendicular to the fabric. Holding the needle parallel to its curve or inserting it at other angles can cause \textit{difficulty driving the needle} through the fabric leading to inverse MPs to correct or re-attempt the gesture.
Many of the ``Grasp/Release(L, Needle)'' in G3 were caused by the left tool grasping the needle trying to adjust how the needle was held while the right tool was trying to drive it through the fabric. This is clinically significant because it can cause excessive strain on the tissue leading to tearing or bleeding. \cite{hutchinson2022analysis} noted that G3 in Suturing had the most ``Multiple attempts'' errors for similar reasons. 
Additionally, all four occurrences of ``Push(Needle, Fabric) Pull(R, Needle)'' were in G3 of the Suturing task where the needle had to be removed from the fabric since poor orientation of the needle made it too difficult to complete the throw as shown in Figure \ref{fig:inv_pair_err}.

\begin{table}[t!]
    \centering
    \caption{Number and percentage of gesture clips with one or more inverse MPs in each gesture over the total number of instances of that gesture.}
    \label{tab:inv_pairs_gesture_clips_gestures}
    \resizebox{\textwidth}{!}{
    \begin{tabular}{l|ccccccccc}
    \toprule
        Task & G2 & G3 & G4 & G6 & G8 & G12 & G13 & G14 & G15 \\
        \midrule
        Suturing & \multicolumn{1}{c}{\begin{tabular}[c]{@{}c@{}}12/166\\ (7.2\%)\end{tabular}} & \multicolumn{1}{c}{\begin{tabular}[c]{@{}c@{}}21/164\\ (12.8\%)\end{tabular}} & \multicolumn{1}{c}{\begin{tabular}[c]{@{}c@{}}7/119\\ (5.9\%)\end{tabular}} & \multicolumn{1}{c}{\begin{tabular}[c]{@{}c@{}}17/163\\ (10.4\%)\end{tabular}} & \multicolumn{1}{c}{\begin{tabular}[c]{@{}c@{}}13/48\\ (27.1\%)\end{tabular}} & \multicolumn{1}{c}{\begin{tabular}[c]{@{}c@{}}\\ \end{tabular}} & \multicolumn{1}{c}{\begin{tabular}[c]{@{}c@{}}\\ \end{tabular}} & \multicolumn{1}{c}{\begin{tabular}[c]{@{}c@{}}\\ \end{tabular}} & \multicolumn{1}{c}{\begin{tabular}[c]{@{}c@{}}\\ \end{tabular}} \\

        Needle Passing & \multicolumn{1}{c}{\begin{tabular}[c]{@{}c@{}}47/117\\ (40.2\%)\end{tabular}} & \multicolumn{1}{c}{\begin{tabular}[c]{@{}c@{}}10/111\\ (9.0\%)\end{tabular}} & \multicolumn{1}{c}{\begin{tabular}[c]{@{}c@{}}7/83\\ (8.4\%)\end{tabular}} & \multicolumn{1}{c}{\begin{tabular}[c]{@{}c@{}}12/112\\ (10.7\%)\end{tabular}} & \multicolumn{1}{c}{\begin{tabular}[c]{@{}c@{}}4/28\\ (14.3\%)\end{tabular}} & \multicolumn{1}{c}{\begin{tabular}[c]{@{}c@{}}\\ \end{tabular}} & \multicolumn{1}{c}{\begin{tabular}[c]{@{}c@{}}\\ \end{tabular}} & \multicolumn{1}{c}{\begin{tabular}[c]{@{}c@{}}\\ \end{tabular}} & \multicolumn{1}{c}{\begin{tabular}[c]{@{}c@{}}\\ \end{tabular}} \\

        Knot Tying & \multicolumn{1}{c}{\begin{tabular}[c]{@{}c@{}}\\ \end{tabular}} & \multicolumn{1}{c}{\begin{tabular}[c]{@{}c@{}}\\ \end{tabular}} & \multicolumn{1}{c}{\begin{tabular}[c]{@{}c@{}}\\ \end{tabular}} & \multicolumn{1}{c}{\begin{tabular}[c]{@{}c@{}}\\ \end{tabular}} & \multicolumn{1}{c}{\begin{tabular}[c]{@{}c@{}}\\ \end{tabular}} & \multicolumn{1}{c}{\begin{tabular}[c]{@{}c@{}}22/70\\ (31.4\%)\end{tabular}} & \multicolumn{1}{c}{\begin{tabular}[c]{@{}c@{}}1/75\\ (1.3\%)\end{tabular}} & \multicolumn{1}{c}{\begin{tabular}[c]{@{}c@{}}17/98\\ (17.3\%)\end{tabular}} & \multicolumn{1}{c}{\begin{tabular}[c]{@{}c@{}}21/73\\ (28.8\%)\end{tabular}} \\

        \midrule

        Total & \multicolumn{1}{c}{\begin{tabular}[c]{@{}c@{}}59/283\\ (20.8\%)\end{tabular}} & \multicolumn{1}{c}{\begin{tabular}[c]{@{}c@{}}31/275\\ (11.3\%)\end{tabular}} & \multicolumn{1}{c}{\begin{tabular}[c]{@{}c@{}}14/202\\ (6.9\%)\end{tabular}} & \multicolumn{1}{c}{\begin{tabular}[c]{@{}c@{}}29/275\\ (10.5\%)\end{tabular}} & \multicolumn{1}{c}{\begin{tabular}[c]{@{}c@{}}17/76\\ (22.4\%)\end{tabular}} & \multicolumn{1}{c}{\begin{tabular}[c]{@{}c@{}}22/70\\ (31.4\%)\end{tabular}} & \multicolumn{1}{c}{\begin{tabular}[c]{@{}c@{}}1/75\\ (1.3\%)\end{tabular}} & \multicolumn{1}{c}{\begin{tabular}[c]{@{}c@{}}17/98\\ (17.3\%)\end{tabular}} & \multicolumn{1}{c}{\begin{tabular}[c]{@{}c@{}}21/73\\ (28.8\%)\end{tabular}} \\

    \bottomrule
    \end{tabular}
    }
\end{table}

In Needle Passing, all three expertise levels \textit{struggled with accurately positioning the needle} because the eye of the ring was a small target leading to a high occurrence of ``Touch/Untouch(Needle, Fabric/Ring)'' in G2 as shown in Table \ref{tab:inv_pairs}. These were a large contributor to the 40.2\% of G2 instances in Needle Passing that contained inverse MPs from Table \ref{tab:inv_pairs_gesture_clips_gestures} and is also consistent with the results from \cite{hutchinson2022analysis} where G2 had the most ``Multiple attempts'' errors in Needle Passing.

When tying knots, the left tool must grasp the suture at the appropriate distance from the knot to make a properly sized wrap, or it will be \textit{difficult for the right tool to grasp and pull the end of the left suture} through the wrap made in G14.
The right tool should grasp the other end of the suture near the end, otherwise it might require more than one re-grasp and pull to get all of it through the knot like in G15 of Knot Tying. Grasping the suture too far away from the knot will necessitate one or both tools re-grasping it closer to the knot and performing a second pull to fully tighten the knot. This was responsible for many of the ``Grasp/Release(R, Thread)'' and ``Grasp/Release(L, Thread)'' which were present in most of the 28.8\% of G15 instances that had inverse MPs in Table \ref{tab:inv_pairs_gesture_clips_gestures}. 

Inverse MPs also occur in additional exchanges of the needle or suture that are used to manipulate the object's position in the tools that could have been accomplished with fewer MPs. 
For example, \textit{correctly orienting the needle} in G8 of Suturing could be accomplished with a single exchange, but Table \ref{tab:inv_pairs_gesture_clips_gestures} shows that this gesture has the highest percentage of gestures with inverse MPs. Many of these were ``Grasp/Release(L, Needle)'' with most from novices. 
Also, G12 can be efficiently performed by grasping the suture in the right tool and passing it to the left as in the surgeon-defined MP sequence shown in Table \ref{tab:gestures_and_mps}. But, a significant number of ``Grasp/Release(L, Thread)'' were mainly from novices and accounted for most of the 31.4\% of G12 instances with inverse MPs in Table \ref{tab:inv_pairs_gesture_clips_gestures}.

\begin{figure}
    \centering
    \includegraphics[page=2, trim=0.25in 5.25in 6.25in 0.25in, clip, width=\textwidth]{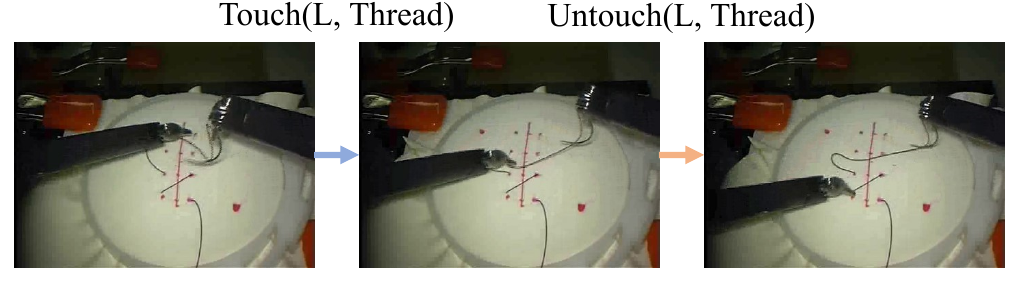}
    \caption{Example of a Touch(L, Thread) Untouch(L, Thread) inverse MP in Suturing.}
    \label{fig:inv_pair_eff}
\end{figure}

\textbf{Adjustment of the environment} was another cause 
of inverse MPs. For example, inverse MPs were used to move or hold the suture out of the way, as shown in Figure \ref{fig:inv_pair_eff}, or to hold the ring in place when positioning the needle in G2 of Suturing and Needle Passing. Specifically, all seven ``Touch/Untouch(L, Thread)'' occurrences were performed by novices in the Suturing task while trying to move the suture out of the way when positioning the tip of the needle at the dot on the fabric. But, they either moved the tool away from the suture or the suture slipped out from beneath the tool. 

It is also interesting to note that there was only one inverse MP in G13. Wrapping the suture around the right grasper was a difficult motion represented by ``Pull(L, Thread)'', but its inverse where the suture became unwrapped was represented by the same MP, and thus was not detected as an inverse MP.
In the manual review, several clips of G14 also showed instances where the wrap slipped off of the right tool, but these contained different types of inverse MPs that either contributed to the suture coming unwrapped or tried to re-wrap the suture.

\begin{table}[]
    \centering
    \caption{Number and percentage of gesture clips with one or more inverse MPs in each gesture over the total number of instances of that gesture by experience level.}
    \label{tab:inv_pairs_gesture_clips_NIE}
    \begin{tabular}{l|ccc|c}
    \toprule
        Task & Novice & Intermediate & Expert & Total \\
        \midrule
        Suturing & \multicolumn{1}{c}{\begin{tabular}[c]{@{}c@{}}45/342\\ (13.2\%)\end{tabular}} & \multicolumn{1}{c}{\begin{tabular}[c]{@{}c@{}}13/161\\ (8.1\%)\end{tabular}} & \multicolumn{1}{c}{\begin{tabular}[c]{@{}c@{}}12/157\\ (7.6\%)\end{tabular}} & \multicolumn{1}{|c}{\begin{tabular}[c]{@{}c@{}}70/660\\ (10.6\%)\end{tabular}} \\

        Needle Passing & \multicolumn{1}{c}{\begin{tabular}[c]{@{}c@{}}27/160\\ (16.9\%)\end{tabular}} & \multicolumn{1}{c}{\begin{tabular}[c]{@{}c@{}}22/138\\ (15.9\%)\end{tabular}} & \multicolumn{1}{c}{\begin{tabular}[c]{@{}c@{}}31/153\\ (20.3\%)\end{tabular}} & \multicolumn{1}{|c}{\begin{tabular}[c]{@{}c@{}}80/451\\ (17.7\%)\end{tabular}} \\

        Knot Tying & \multicolumn{1}{c}{\begin{tabular}[c]{@{}c@{}}39/139\\ (28.1\%)\end{tabular}} & \multicolumn{1}{c}{\begin{tabular}[c]{@{}c@{}}14/87\\ (16.1\%) \end{tabular}} & \multicolumn{1}{c}{\begin{tabular}[c]{@{}c@{}}7/90\\ (7.8\%) \end{tabular}} & \multicolumn{1}{|c}{\begin{tabular}[c]{@{}c@{}}61/316\\ (19.3\%)\end{tabular}} \\

        \midrule

        Total & \multicolumn{1}{c}{\begin{tabular}[c]{@{}c@{}}111/641\\ (17.3\%)\end{tabular}} & \multicolumn{1}{c}{\begin{tabular}[c]{@{}c@{}}50/386\\ (13.0\%)\end{tabular}} & \multicolumn{1}{c}{\begin{tabular}[c]{@{}c@{}}50/400\\ (12.5\%)\end{tabular}} & \multicolumn{1}{|c}{\begin{tabular}[c]{@{}c@{}}211/1427\\ (14.8\%)\end{tabular}} \\

    \bottomrule
    \end{tabular}
\end{table}

\subsection{Relationship between Inverse MPs and Skill Level}

\textbf{Occurrences of Inverse MPs by Expertise Level:}
Table \ref{tab:inv_pairs_gesture_clips_NIE} shows the number and percentage of gesture clips with one or more inverse MPs over the total number of clips by expertise level. When considering expertise level, we use percentages due to the data imbalance of having four novices, two intermediates, and two experts. 
Overall, and for the Suturing and Knot Tying tasks, \textit{the percentage of gestures with inverse MPs decreased with increasing expertise level.} However, this trend is not seen in the Needle Passing task where experts had the highest percentage of gestures with inverse MPs. This could be because Needle Passing is a more difficult task than Suturing since greater accuracy is required to thread the needle through the eye of the ring. This also makes it less surgically realistic and we found that more than half of the inverse MPs in the Needle Passing task were ``Touch/Untouch(Needle, Ring)'' from experts which was even greater than the number from novices. 

\noindent\textbf{Correlation between Inverse MPs and GRS Scores:}
Tables \ref{tab:rho_counts} and \ref{tab:rho_durations} show Spearman's correlation ($\rho$) between the number and total duration of inverse MPs in a trial with the GRS scores and subscores. There are significant and strong negative correlations for all GRS scores and subscores in the Suturing and Knot Tying tasks with $\rho < -0.5$ except for the ``Suture/needle handling'' subscore in Knot Tying which has correlations of -0.46 and -0.49. This means that a \textit{greater number of occurrences and longer durations of inverse MPs can be indicative of lower skill}. However, in the Needle Passing task, the correlations are only between -0.06 and -0.33 with larger p-values indicating a weaker and not significant correlation.

\begin{table}[h!]
    \vspace{-1em}
    \centering
    \caption{Correlation ($ \rho $) between GRS subscores and the total number of inverse MPs.}
    \label{tab:rho_counts}
    \begin{tabular}{lccccccc}
    \toprule
         & \multicolumn{2}{c}{Suturing} & \multicolumn{2}{c}{Needle Passing} & \multicolumn{2}{c}{Knot Tying} \\
        \cmidrule(lr){2-3} \cmidrule(lr){4-5} \cmidrule(lr){6-7}
        GRS Subscore & $ \rho $ & p-value & $ \rho $ & p-value & $ \rho $ & p-value \\
        \midrule
        Respect for tissue & -0.63 & $ < $0.001 & -0.13 & 0.517 & -0.63 & $ < $0.001 \\
        Suture/needle handling & -0.52 & $ < $0.001 & -0.15 & 0.432 & -0.46 & 0.005 \\
        Time and motion & -0.64 & $ < $0.001 & \textbf{-0.20} & 0.296 & -0.64 & $ < $0.001 \\
        Flow of operation & -0.58 & $ < $0.001 & -0.06 & 0.775 & \textbf{-0.65} & $ < $0.001 \\
        Overall performance & \textbf{-0.65} & $ < $0.001 & -0.14 & 0.471 & -0.61 & $ < $0.001 \\
        Quality of final product & -0.50 & 0.001 & -0.03 & 0.890 & -0.59 & $ < $0.001 \\
        \midrule
        GRS Score & \textbf{-0.65} & $ < $0.001 & -0.19 & 0.338 & -0.63 & $ < $0.001 \\
    \bottomrule
    \end{tabular}
\end{table}

\begin{table}[b!]
    \vspace{-2em}
    \centering
    \caption{Correlation ($ \rho $) between GRS subscores and total duration of inverse MPs.}
    \label{tab:rho_durations}
    \begin{tabular}{lccccccc}
    \toprule
         & \multicolumn{2}{c}{Suturing} & \multicolumn{2}{c}{Needle Passing} & \multicolumn{2}{c}{Knot Tying} \\
        \cmidrule(lr){2-3} \cmidrule(lr){4-5} \cmidrule(lr){6-7}
        GRS Subscore & $ \rho $ & p-value & $ \rho $ & p-value & $ \rho $ & p-value \\
        \midrule
        Respect for tissue & -0.59 & $ < $0.001 & -0.26 & 0.188 & -0.58 & $ < $0.001 \\
        Suture/needle handling & -0.61 & $ < $0.001 & -0.28 & 0.151 & -0.49 & 0.002 \\
        Time and motion & \textbf{-0.71} & $ < $0.001 & -0.31 & 0.113 & \textbf{-0.71} & $ < $0.001 \\
        Flow of operation & -0.68 & $ < $0.001 & -0.20 & 0.305 & -0.66 & $ < $0.001 \\
        Overall performance & -0.66 & $ < $0.001 & -0.32 & 0.100 & -0.63 & $ < $0.001 \\
        Quality of final product & -0.57 & $ < $0.001 & -0.22 & 0.258 & -0.61 & $ < $0.001 \\
        \midrule
        GRS Score & \textbf{-0.71} & $ < $0.001 & \textbf{-0.33} & 0.087 & -0.65 & $ < $0.001 \\
    \bottomrule
    \end{tabular}
\end{table}

\begin{figure}
    \centering
    \begin{subfigure}{\textwidth}
        \centering
        \includegraphics[trim = 0.4in 2.5in 0.25in 0.25in, clip, width=\textwidth]{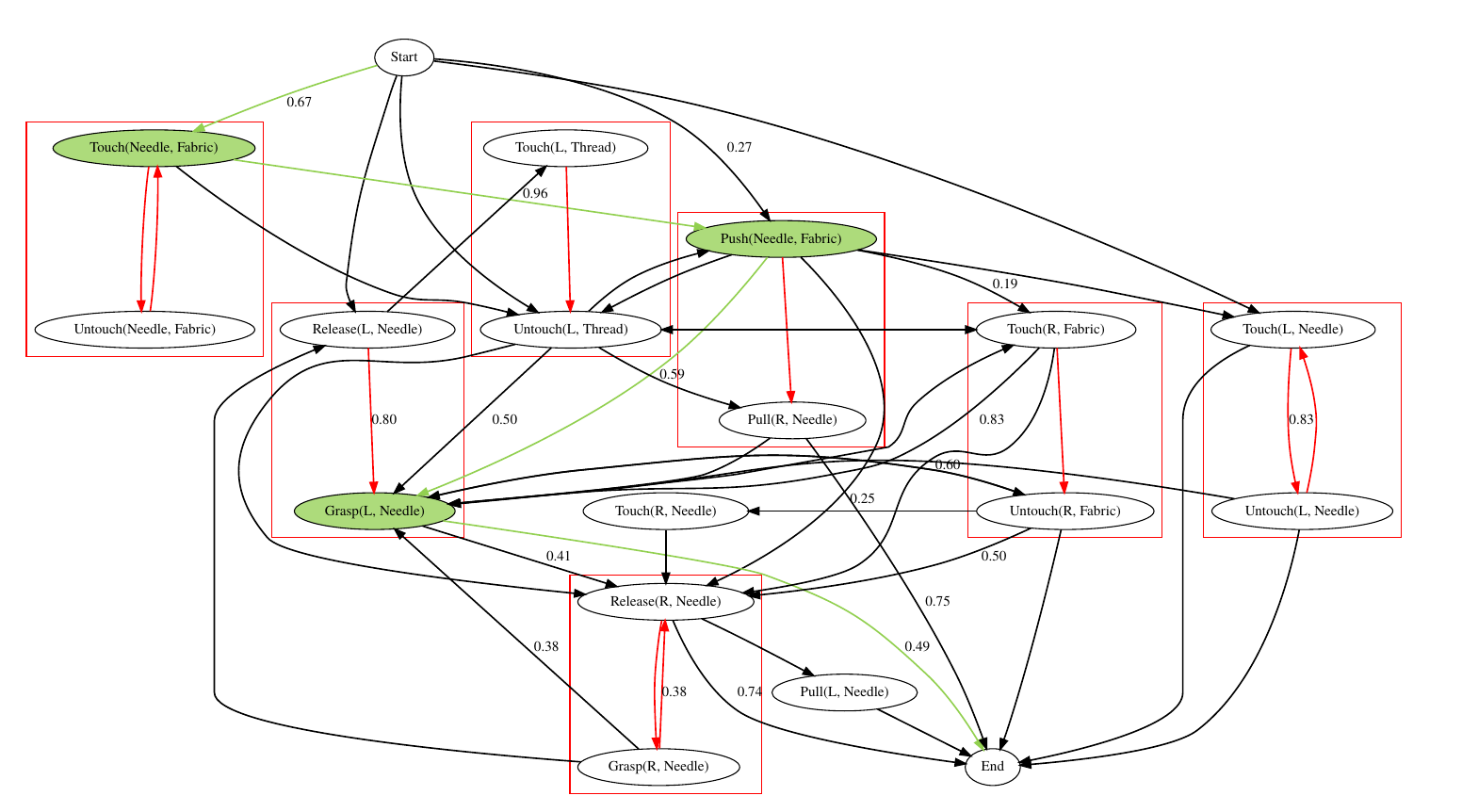}
        \caption{Novice}
        \label{fig:S_G3_N}
    \end{subfigure}
    \vspace{-1em}
    \begin{subfigure}{0.49\textwidth}
        \centering
        \includegraphics[trim = 0.3in 0.75in 1.5in 0.3in, clip, width=\textwidth]{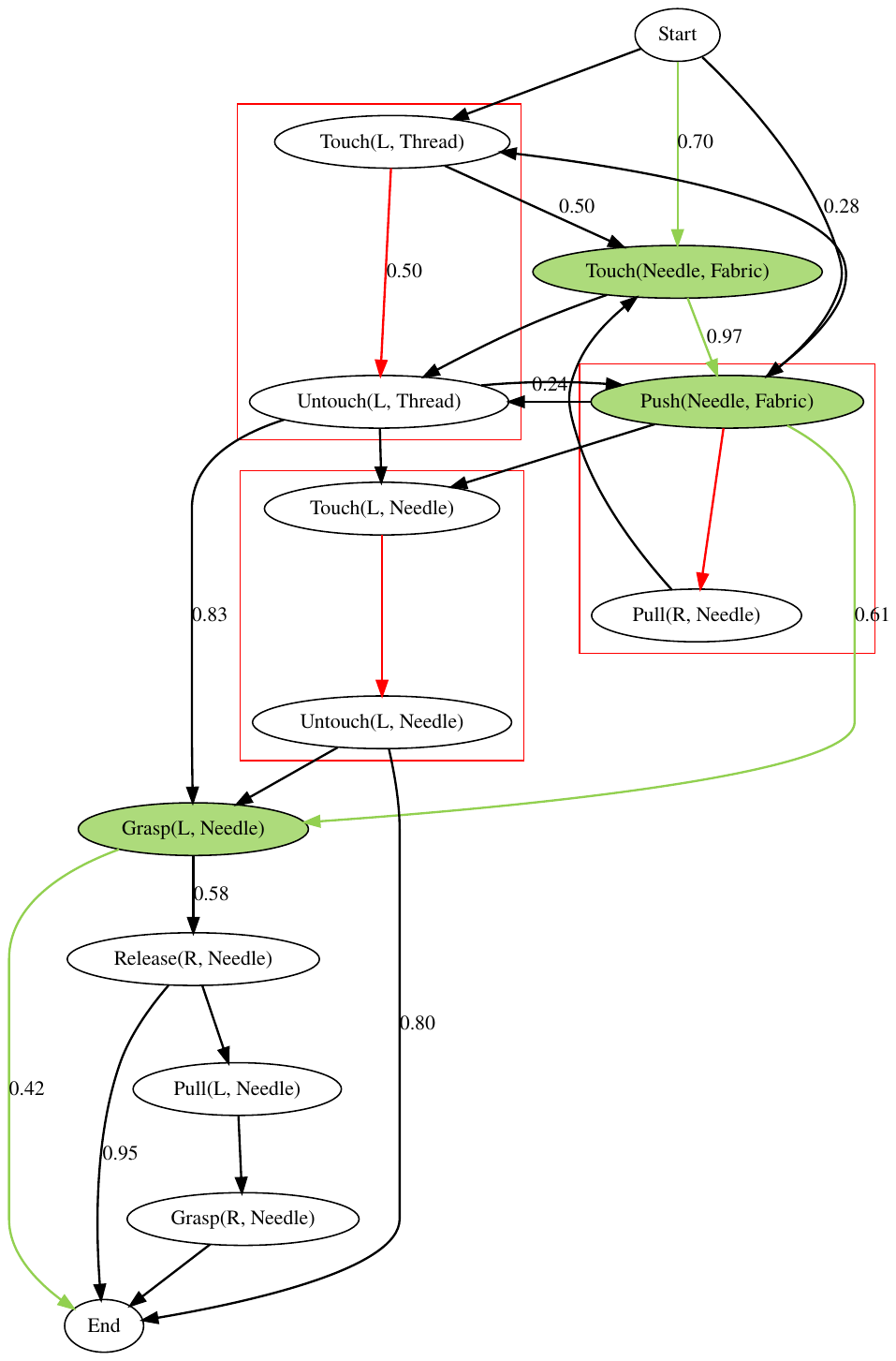}
        \caption{Intermediate}
        \label{fig:S_G3_I}
    \end{subfigure}
    \hfill
    \begin{subfigure}{0.49\textwidth}
        \centering
        \includegraphics[trim = 0.3in 0.65in 1.25in 0.3in, clip, width=\textwidth]{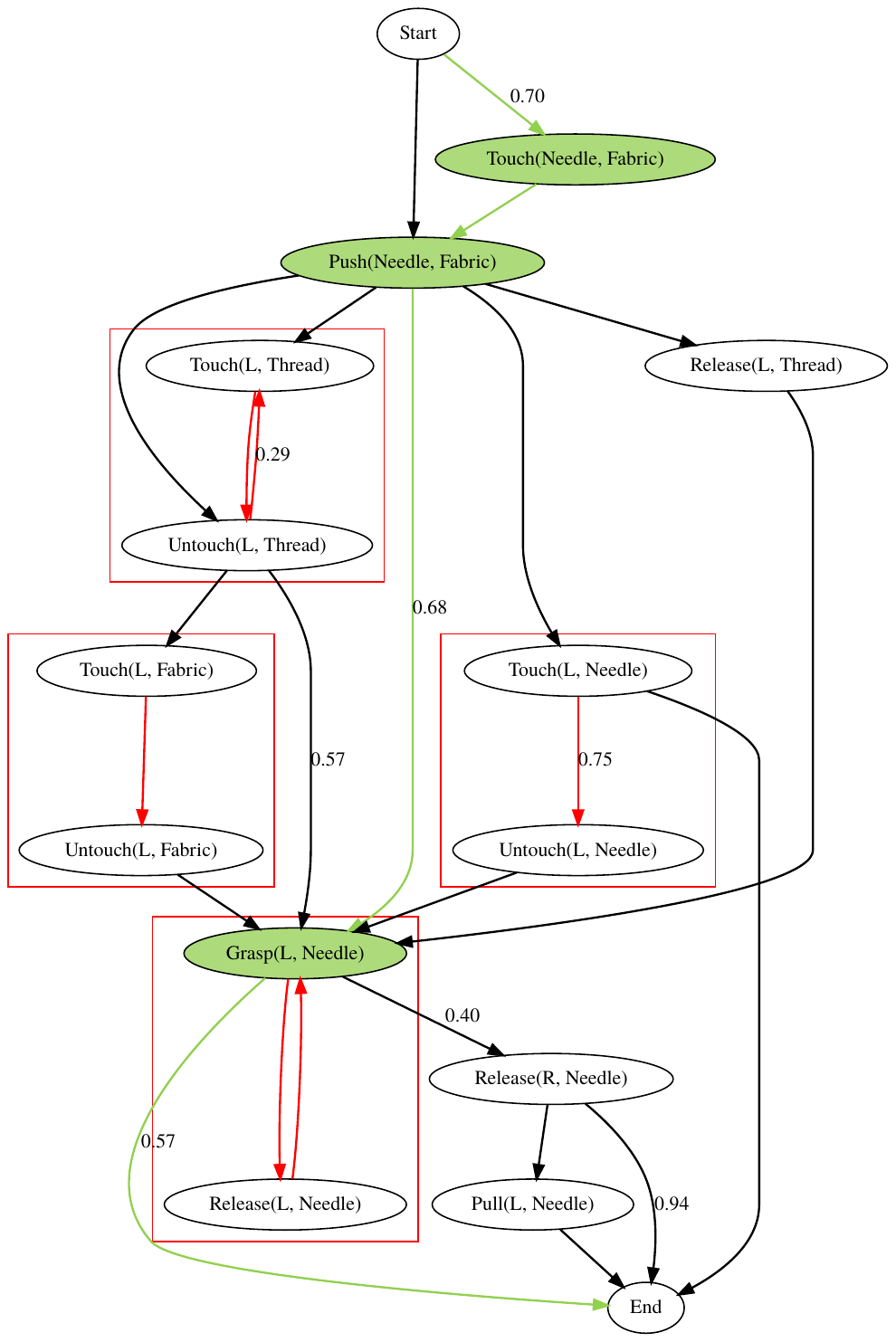}
        \caption{Expert}
        \label{fig:S_G3_E}
    \end{subfigure}
    
    \caption{State transition diagrams for Suturing G3 as performed by \ref{fig:S_G3_N} novices, \ref{fig:S_G3_I} intermediates, and \ref{fig:S_G3_E} experts with the surgeon-defined MP sequence in green and inverse MPs in red.}
    \label{fig:S_G3_graphs}
\end{figure}

\noindent\textbf{MP State Graphs for Expertise Levels:}
State transition diagrams for each gesture, like Figure \ref{fig:S_G3_graphs} for G3 in Suturing, visualize the MP sequences performed by different experience levels. The surgeon-defined sequences are part of the dominant path in each graph, but the graphs showed variation with task and experience level. In Suturing and Knot Tying, \textit{novices tended to have more states and more complicated graphs than experts and intermediates} similar to \cite{reiley2009task}, but in Needle Passing, experts usually had more complicated graphs than novices.

Our observations are similar to \cite{varadarajan2009data} where G3 was modeled with five dexemes, including dexemes a, b, and c showing the right tool driving the needle which aligns with Push(Needle, Fabric) and Dexemes d and e, showing the left tool reaching and grasping the needle, which aligns with Touch(L, Needle) and Grasp(L, Needle), respectively. This agrees with the surgeon-defined MP sequence for G3 of Suturing in Table \ref{tab:gestures_and_mps}.

\section{Discussion and Conclusion}
\label{sec:conclusion}

We analyze surgical activity at the gesture and motion levels and find that the sequence of MPs could help detect labeling errors in surgical gestures. 
Inverse MPs are defined and identified as
motion-level patterns often used as recovery actions to correct the position or orientation of objects. Although inverse MPs are sometimes required to perform gestures, we find that their number and duration 
strongly correlates with lower GRS scores for two dry-lab surgical tasks. 
Recently, \cite{hendricks2023exploring} found minimal correlation between self-proclaimed expertise and GRS scores in the JIGSAWS dataset, but subjects with higher GRS scores performed certain tasks with greater similarity.
Thus, future work could investigate how the MPs used to perform gestures may vary with GRS scores.

This study could enable developing a pipeline for interpretable surgical skill assessment that can automatically identify inverse MPs within the MP sequences generated by an action recognition model~\cite{hutchinson2023evaluating}. Showing video clips of these inefficient or problematic motions will provide specific explanations of the causes of lower performance scores. However, better performing action recognition models are needed to obtain reliable and precise motion sequences and support this method of skill assessment.


\backmatter





\bmhead{Acknowledgments}

This work was supported in part by the National Science Foundation grants DGE-1842490 and CNS-2146295. We thank Dr. Schenkman and Dr. Cantrell for their medical feedback.

\section*{Declarations}



\subsection*{Competing Interests}
The authors declare that they have no conflict of interest.
\subsection*{Ethics Approval and Informed Consent}
This article does not contain any studies involving human participants performed by any of the authors.

\noindent

\bibliographystyle{plain}
\bibliography{main.bib}

\end{document}